\documentclass[sigconf]{acmart}
\usepackage[linesnumbered,ruled,vlined]{algorithm2e}
\usepackage{hyperref}
\usepackage{amsmath}
\usepackage{titlesec}

\titleformat{\subsection}[runin]
  {\normalfont\Large\bfseries}{\thesubsection}{1em}{}
\titlespacing*{\subsection}{0pt}{0.5em}{1em}

\AtBeginDocument{%
  \providecommand\BibTeX{{%
    \normalfont B\kern-0.5em{\scshape i\kern-0.25em b}\kern-0.8em\TeX}}}

\setcopyright{acmlicensed}
\copyrightyear{2024}
\acmYear{2024}
\acmDOI{XXXXXXX.XXXXXXX}

\acmConference[Conference acronym 'XX]{Make sure to enter the correct
  conference title from your rights confirmation email}{June 03--05,
  2018}{Woodstock, NY}
\acmISBN{978-1-4503-XXXX-X/18/06}

\graphicspath{{./images/}}

\begin{document}


\title{MEDFuse: Multimodal EHR Data Fusion with Masked Lab-Test Modeling and Large Language Models}

\author{Thao Minh Nguyen Phan}
\authornote{Both authors contributed equally to this research.}

\author{Cong-Tinh Dao}
\authornotemark[1]
\affiliation{%
  \institution{National Yang Ming Chiao Tung University}
  \city{Hsinchu}
  \country{Taiwan}
}
\email{{pnmthaoct, dcongtinh}@gmail.com}

\author{Chenwei Wu}
\affiliation{%
  \institution{University of Michigan}
  \state{Michigan}
  \country{USA}}
\email{chenweiw@umich.edu}

\author{Jian-Zhe Wang}
\affiliation{%
  \institution{National Yang Ming Chiao Tung University}
  \city{Hsinchu}
  \country{Taiwan}}
\email{jzwang.cs09@nycu.edu.tw}

\author{Shun Liu}
\affiliation{%
  \institution{Shanghai University of Finance and Ecomonics}
  \city{Shanghai}
  \country{China}}
\email{kevinliuleo@gmail.com}

\author{Jun-En Ding}
\affiliation{%
  \institution{Stevens Institute of Technology Hoboken}
  \state{New Jersey}
  \country{USA}}
\email{jding17@stevens.edu}

\author{David Restrepo}
\affiliation{%
  \institution{Massachusetts Institute of Technology}
  \state{Massachusetts}
  \country{USA}}
\email{davidres@mit.edu}

\author{Feng Liu}
\affiliation{%
  \institution{Stevens Institute of Technology Hoboken}
  \state{New Jersey}
  \country{USA}}
\email{fliu22@stevens.edu}

\author{Fang-Ming Hung}
\affiliation{%
  \institution{Far Eastern Memorial Hospital}
  \city{New Taipei}
  \country{Taiwan}}
\email{philip@mail.femh.org.tw}

\author{Wen-Chih Peng}
\affiliation{%
  \institution{National Yang Ming Chiao Tung University}
  \city{Hsinchu}
  \country{Taiwan}}
\email{wcpengcs@nycu.edu.tw}

\renewcommand{\shortauthors}{Phan et al.}

\begin{abstract}
Electronic health records (EHRs) are multimodal by nature, consisting of structured tabular features like lab tests and unstructured clinical notes. In real-life clinical practice, doctors use complementary multimodal EHR data sources to get a clearer picture of patients' health and support clinical decision-making. However, most EHR predictive models do not reflect these procedures, as they either focus on a single modality or overlook the inter-modality interactions/redundancy. In this work, we propose \textbf{MEDFuse}, a \textbf{M}ultimodal \textbf{E}HR \textbf{D}ata \textbf{F}usion framework that incorporates masked lab-test modeling and large language models (LLMs) to effectively integrate structured and unstructured medical data. MEDFuse leverages multimodal embeddings extracted from two sources: LLMs fine-tuned on free clinical text and masked tabular transformers trained on structured lab test results. We design a disentangled transformer module, optimized by a mutual information loss to 1) decouple modality-specific and modality-shared information and 2) extract useful joint representation from the noise and redundancy present in clinical notes. Through comprehensive validation on the public MIMIC-III dataset and the in-house FEMH dataset, MEDFuse demonstrates great potential in advancing clinical predictions, achieving over 90\% F1 score in the 10-disease multi-label classification task. 

\end{abstract}




\received{20 February 2024}  
\received[revised]{12 March 2024} 
\received[accepted]{5 June 2024}  

\maketitle

\section{Introduction}

Electronic Health Records (EHRs) are widely adopted in healthcare, documenting a wealth of heterogeneous patient data comprised of tabular records and unstructured clinical notes. Tabular records encompass essential medical concepts such as diagnoses, medications, and laboratory test results, providing a structured overview of a patient's health. In contrast, clinical notes are extensive, free-text documents written by healthcare providers, offering a more detailed and nuanced account of the patient's history, clinical findings, and progress. The vast volume and diversity of multimodal data within EHRs present a unique opportunity for deep learning technologies to improve the prediction and management of diseases \cite{ding2024large,restrepo2024df}.
 Nevertheless, the heterogeneous nature and large quantity of redundancy in multimodal EHR inputs pose significant challenges for medical AI practitioners to effectively distill and fuse clinically meaningful information for disease prediction.


The primary question at hand is: \textit{can we effectively obtain and integrate useful representations for different EHR modalities to improve clinical predictions?} Current research in deep EHR modeling \cite{luo2020hitanet, ma2017dipole} often focuses on single data modalities, often neglecting the integration of significant insights from unstructured medical notes and lab tests. This oversight can limit the model from learning a more comprehensive view of patient health conditions. Lab tests consist of high-dimensional, usually discrete tabular data; however, the conventional approach models structured EHR data as numerical vectors, overlooking complex interactions between
individual variables and does not consider their interactions \cite{choi2016retain,li2020behrt, zhang2020inprem,luo2020hitanet, ma2017dipole}. More recent work has moved towards deep learning architectures like Bert and LLMs. LLMs fine-tuned on clinical data have shown promise in unstructured clinical notes in understanding tasks like answering medical questions and making few-shot predictions \cite{thirunavukarasu2023large}. However, there is a large body of evidence showing that LLMs are still having a hard time capturing the nuances of numerical lab test data and are underperforming
on tabular prediction tasks \cite{grinsztajn2022tree, bellamy2023labrador, hegselmann2023tabllm}.

Another significant challenge in fusing information from different types of EHR data is \textit{How do we distill the overlapping clinically important features from both modalities?} The information contained within different modalities can be categorized as either modality-specific or modality-shared \cite{liang2023quantifying}. For example, a patient's dietary habits would be considered information specific to the clinical notes modality; hypertension record and lab test value would be regarded as modality-shared information. Existing efforts like multimodal EHR contrastive learning \cite{cai2024contrastive} have primarily focused on integrating the modality-shared information by emphasizing the inherent consistency through alignment techniques. However, this approach often leads to the common information dominating the alignment and integration process, resulting in the distinctive perspectives offered by each modality being disregarded. Lab tests and clinical notes also possess highly different noise-to-information ratios, making it hard to distill useful joint
representation from the noises and redundancy present in EHR. Therefore, there is an urgent need for methods to extract the diverse yet collaborative perspectives both modalities offer for informing therapeutic decision-making. 

In this work, we propose MEDFuse, a novel Multimodal EHR Data Fusion diagnostic model consisting of modality-specific embedding extractors followed by a disentangled transformer for multimodal fusion. Our model integrates embeddings between fine-tuned LLMs on unstructured clinical text and masked lab-test modeling models pre-trained on structured laboratory results. We further utilize a disentangled transformer optimized by mutual information loss to decouple modality-specific and modality-common information and learn meaningful joint representations for downstream prediction tasks. The key contributions of our work are as follows:
\vspace{-3mm}
\begin{itemize}
    \item We propose a novel diagnostic model integrating structured lab test data and unstructured clinical notes, utilizing embeddings from fine-tuned LLMs and Masked Lab-Test Modeling, enhancing understanding of diverse clinical information.
    
    \item We improved joint patient representation by incorporating a disentangled transformer module to effectively separate and integrate modality-specific and shared information, leading to better prediction outcomes across multiple diseases.

    \item We conducted empirical evaluations to illustrate our model’s effectiveness through EHR datasets on various metrics.
\end{itemize}

\section{Related work}
\subsection{EHR For Multi-Label Disease Prediction}
Most recent works in medical Multi-Label Text Classification (MLTC) entirely rely on medical texts. For instance, Kim et al. \cite{chen2015convolutional} introduced a convolutional attention network designed to extract meaningful document representations across varying text lengths. Recent developments in LLMs, such as those discussed by Luo et al. (2022) and Elliot et al. \cite{bolton2024biomedlm}, utilize extensive data from medical literature for domain-specific tasks such as natural language inferencing. Additionally, some studies have employed graph neural networks (GNNs) to organize sequences from Electronic Medical Records (EMR) into hierarchical graphs \cite{wu2021counterfactual}, or to integrate entity relationships from text using attention mechanisms in neural networks \cite{chen2019deep, dun2021kan}. Nevertheless, many of these studies overlook the potential advantages of integrating medical expert knowledge from official guidelines and critical blood tests. A combined approach that harnesses both unstructured and structured data could offer extra help to offset issues like label and data scarcity in the medical domain.
\subsection{Extraction of Clinical Relevant Information from Multimodal EHR}

Recent work has leveraged self-supervised learning methods, like contrastive pretraining of clinical notes \cite{cai2024contrastive} and prompt-based large language modeling \cite{ding2024large,hegselmann2023tabllm}, to facilitate multimodal learning of EHR data. The former encourages the alignment between paired patient data via contrastive loss, and the latter usually directly converts the structured data into text by prompt templates and feeds it into LLMs. However, if the data fusion process focuses solely on aligning the common information, such as diabetes history (text) and blood glucose levels (lab), the rich, modality-specific insights like exercise habits may be overlooked. This can limit understanding of the patient's health and impact predictive models and clinical decisions. Therefore, it is essential to develop multimodal EHR data fusion techniques that can effectively capture and integrate both modality-specific and modality-shared information. 

\begin{figure*}[h!]
\centering
\includegraphics[width=0.7\textwidth]{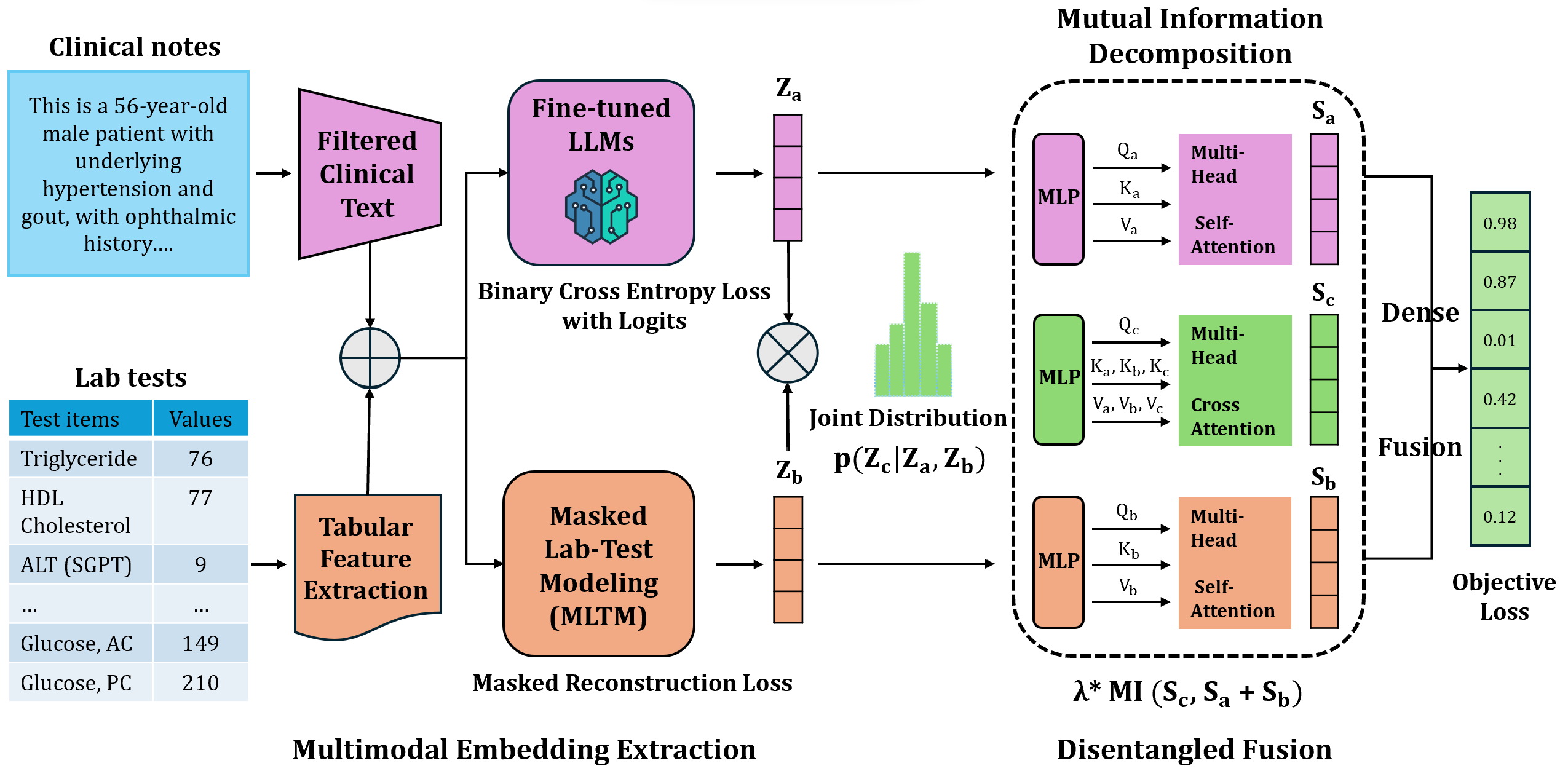}
\vspace{-1mm}
\caption{The proposed model architecture.}
\label{fig:model}
\vspace{-1mm}
\end{figure*}

\section{Method}
\subsection{Overview}

Given the clinical notes and lab tests of the patient’s current and historical visits, MEDFuse integrates clinical notes and lab test data to create a comprehensive patient representation for accurate multi-disease prediction. Firstly, as illustrated in Figure \ref{fig:model}, in the Multimodal Embedding Extraction stage, textual data from clinical notes, including detailed patient information and medical history, are filtered and structured. Simultaneously, abnormal numerical data from various lab tests, such as triglyceride levels, HDL cholesterol, ALT (SGPT), and glucose levels, are extracted and formatted into textual data by prompt templates. The filtered clinical text is then processed by fine-tuned LLMs to generate embeddings that capture its semantic meaning. In parallel, the raw structured tabular data are processed using a domain-specific masked lab-test model to create embeddings representing the quantitative lab data. The two embeddings are then passed through the disentangled transformer module for multimodal fusion and final disease prediction.

\subsection{Multimodal Embedding Extraction}
\vspace{-1.6mm}
\subsubsection{\textbf{Fine-tuning LLMs on Unstructured Text}} Clinical notes, comprising diverse fields derived from physicians' diagnoses, form the textual component of the dataset. We filtered the text by Chief Complaint, Present Illness, Medical History, and Medication on Admission. These specific fields are crucial for accurately predicting a patient's disease. To integrate this tabular data with the textual clinical notes, we converted the tabular data into a textual format, a process referred to as tabular feature extraction. This method involves extracting abnormal lab test results and formatting them into a text template — ``These are abnormal results recorded: ITEMID <ITEMID>: <VALUE> <VALUEUOM>; ITEMID <ITEMID>: <VALUE> <VALUEUOM>; ...;''. Here, <ITEMID> refers to the specific lab test names, <VALUE> indicates the test values, and <VALUEUOM> denotes the units of measure for the test values.

Inspired by the recent success of fine-tuning the language models \cite{devlin2018bert,liu2019roberta} for classification purposes, we fine-tuned various LLMs for disease prediction. Our best-performing backbone is the publicly accessible Medical-Llama3-8B model \cite{medicalllama3modelcard}, which is fine-tuned from Meta-Llama-3-8B \cite{llama3modelcard}. It is trained on a comprehensive medical chatbot dataset and optimized for addressing health-related inquiries. We extracted latent vector representations from the final layer of the Llama decoder, which was originally engineered for autoregressive prediction of subsequent tokens. These extracted vectors were subsequently processed through feed-forward neural layers, effectively transforming them into a label space. The output from these transformations, in the form of logits, was then utilized to perform discriminative classification based on labels. This method aims to harness the latent embedding of LLMs to achieve targeted, efficient task adaptation.
\subsubsection{\textbf{Masked Lab-Test Modeling}}
The Masked Lab-Test Modeling (MLTM) module extends the Masked Autoencoders (MAE) \cite{he2022masked,song2023zeroprompt,chen2024tokenunify} framework to reconstruct masked components based on observed components in EHR data. MLTM consists of a encoder that maps observed values to their representations and a decoder that reconstructs the masked values from the latent representations. To account for the inherent incompleteness in the imputation task, MLTM employs an additional masking approach in the training to make sure a uniform 75\% value is masked out. The encoder applies a learnable linear encoding function $wx + b$ to each unmasked $x$ and passes through a transformer architecture, while the decoder operates on the embeddings of both observed and masked values. Positional encoding is added to the embeddings to preserve the lab test positions. The reconstruction loss is defined as the mean square error between the reconstructed and original values on the re-masked and unmasked sets. MLTM is designed with an asymmetric architecture, using a deep encoder and a shallow decoder to extract useful lab-test representations.

\subsection{Disentangled Transformer Module}

\begin{table*}
\caption{Training and Validation Performance Comparison of Various Models on the MIMIC-III Dataset.}
\vspace{-3.5mm}
\begin{tabular}{lccccc}
\hline
\textbf{Model} & \textbf{Precision} & \textbf{Recall} & \textbf{F1 macro} & \textbf{F1 weighted} & \textbf{Accuracy} \\
\hline
Bert & 0.8333 / 0.6790 & 0.2000 / 0.2000 & 0.1818 / 0.1618 & 0.3686 / 0.3162 & 0.6515 / 0.2692\\
\hline
LoRA Mistral-7B-v0.1 & 0.8759 / 0.8616 & 0.8459 / 0.8289 & 0.8449 / 0.8274 & 0.9007 / 0.8886 & 0.9089 / 0.8974 \\
\hline
LoRA Llama-2-7B-hf & 0.8828 / 0.8585 & 0.8592 / 0.8364 & 0.8559 / 0.8301 & 0.9097 / 0.8924 & 0.9168 / 0.9004\\
\hline
LoRA Meta-Llama2-13B & 0.9153 / 0.8732 & 0.8852 / 0.8430 & 0.8874 / 0.8414 & 0.9297 / 0.8990 & 0.9363 / 0.9071\\
\hline
LoRA Meta-Llama3-8B & 0.8899 / 0.8667 & 0.8569 / 0.8306 & 0.8579 / 0.8305 & 0.9121 / 0.8935 & 0.9211 / 0.9040 \\
\hline
LoRA Medical-Llama3-8B & 0.9283 / 0.8807 & 0.9008 / 0.8474 & 0.9026 / 0.8466 & 0.9367 / 0.9003 & 0.9417 / 0.9068\\
\hline
\textbf{MEDFuse} & \textbf{0.9375 / 0.9025} & \textbf{0.9217 / 0.8534} & \textbf{0.9216 / 0.8615} & \textbf{0.9462 / 0.9103}& \textbf{0.9535 / 0.9122} \\
\hline
\end{tabular}
\label{tab:mimic}
\vspace{-1.5mm}
\end{table*}

\begin{table*}
\caption{Training and Validation Performance on the FEMH Dataset.}
\vspace{-3.5mm}
\begin{tabular}{lccccc} \hline
\textbf{Model} & \textbf{Precision} & \textbf{Recall} & \textbf{F1 macro} & \textbf{F1 weighted} & \textbf{Accuracy} \\ \hline
LoRA Medical-Llama3-8B & 0.8702 / 0.8691 & 0.8496 / 0.8478 & 0.8453 / 0.8435 & 0.9182 / 0.9167 & 0.9267 / 0.9252 \\ \hline
\textbf{MEDFuse} & \textbf{0.8839 / 0.8823} & \textbf{0.8707 / 0.8670} & \textbf{0.8637 / 0.8607} & \textbf{0.9260 / 0.9243} & \textbf{0.9311 / 0.9296} \\ \hline
\end{tabular}
\label{tab:femh}
\vspace{-1.5mm}
\end{table*}

\begin{table*}
\caption{Ablation Study on Training and Validation Performance on the MIMIC-III Dataset.}
\vspace{-3.5mm}
\begin{tabular}{lccccc}
\hline
\textbf{Model} & \textbf{Precision} & \textbf{Recall} & \textbf{F1 macro} & \textbf{F1 weighted} & \textbf{Accuracy} \\
\hline
MEDFuse w/o (MLTM \& LABTEXT) & 0.8882 / 0.8580 & 0.8663 / 0.8406 &  0.8620 / 0.8321 & 0.9100 / 0.8901 & 0.9148 / 0.8955 \\
\hline
MEDFuse w/o (MLTM \& TEXT) & 0.6553 / 0.6224 & 0.6461 / 0.6203 & 0.6282 / 0.5980 & 0.7869 / 0.7627 & 0.8239 / 0.8008 \\
\hline
MEDFuse w/o TEXT & 0.7730 / 0.7666 & 0.7912 / 0.7923 & 0.7600 / 0.7573 & 0.8331 / 0.8230 & 0.8331 / 0.8271 \\
\hline
MEDFuse w/o Disentangled Transformer & 0.9330 / 0.8974 & 0.9164 / 0.8483 &0.9162 / 0.8564 & 0.9417 / 0.9074 & 0.9489 / 0.9082 \\
\hline
\textbf{MEDFuse} & \textbf{0.9375 / 0.9025} & \textbf{0.9217 / 0.8534} & \textbf{0.9216 / 0.8615} & \textbf{0.9462 / 0.9103}& \textbf{0.9535 / 0.9122} \\
\hline
\end{tabular}
\label{tab:ablation-study}
\vspace{-1.5mm}
\end{table*}

Initially, features from each modality are multiplied by the Kronecker product to approximate a joint distribution, $C = A \bigotimes B \in \mathbb{R}^{(m\times(a)\times(b))}$, effectively capturing the pairwise interactions. Self-attention is applied to $Z_a$ and $Z_b$ to obtain $S_a$ and $S_b$, controlling the expressivity of each modality and preventing noisy features. Subsequently, the common information of the joint distribution is extracted via cross attention of $Q_c$, $K_c+K_a+K_b$, and $V_c+V_a+V_b$ to model modality-common features $S_c$. To preserve modality-specific information, we minimize the Mutual Information (MI) loss between concatenated $S_a+S_b$ and $S_c$. As the computation of mutual information is intractable, we calculate a variational upper bound called contrastive log-ratio upper bound (vCLUB) as an MI estimator to achieve MI minimization. Given two variables $a$ and $b$, the $L_v^{CLUB}(a,b)$ is calculated as follows \cite{zhang2024prototypical}:
\vspace{-1mm}
\begin{align}
L_v^{CLUB}(a, b) &= \mathbb{E}_p(a, b)\left[\log q_\theta (b|a)\right] - \mathbb{E}_p(a)\mathbb{E}_p(b)\left[\log q_\theta (b|a)\right] \nonumber \\
&= \frac{1}{N^2} \sum_{i=1}^{N} \sum_{j=1}^{N} \left[ \log q_\theta (b_i|a_i) - \log q_\theta (b_j|a_i) \right]
\end{align}

We employ an MLP $q_\theta(b|a)$ to provide a variational approximation of $q_\theta(b|a)$, which can be optimized by maximizing the log-likelihood \cite{zhang2024prototypical}:
$L_{\text{estimator}}(a, b) = \frac{1}{N} \sum_{i=1}^{N} \log q_{\theta}(b_i | a_i)$. The MI Loss is then calculated as: 
$\text{MI Loss} = L_v^{CLUB}(S_a + S_b) + L_{\text{estimator}}(S_a + S_b, S_c)$.

After optimizing the mutual information between the modality-specific information and the modality-common information, we utilize dense fusion \cite{holste2023improveddensefusion} to enable denser interaction between modalities. Instead of directly connecting a prediction classifier on top of the fused representation $S_c$, we learn deeper representations of the clinical notes and lab test features and add skip connections to concatenate with the fused representation, forming a final fused embedding:
$h_a=f_a (S_a) \text{ and } h_b=f_b(S_b)$
where $f_a$ and $f_b$ are fully-connected layers. This final representation not only aggregates the modality-specific features but also incorporates the modality-common representation from the previous stage of the network:
$h_{final}=concat(h_a,S_c,h_b)$.
Finally, a dense block $g$ is used to generate $y=g(h_{final})$, and the model is trained by optimizing the prediction loss (focal loss for multilabel prediction). This allows for dense interaction of features from each modality, aggregating information across different stages of the network. The final loss optimizes a combination of the prediction objective and the mutual information loss, controlled by a hyperparameter $\lambda$ with a value range of [0,1]. In this case, we choose a value of 0.1.
$\text{Loss}_{\text{final}} = L_{\text{objective}}(g(h_{\text{final}})) + \lambda * \text{MI}(\text{concat}(S_a, S_b), S_c)$

\section{EXPERIMENTS}

\subsection{Datasets and Metrics}
To evaluate the performance of the methods under comparison, we employed two real-world EHR datasets: MIMIC-III \cite{mimic3} and FEMH. We collected five years of EHRs from the Far Eastern Memorial Hospital (FEMH) in Taiwan from 2017 to 2021. The dataset includes 1,420,596 clinical notes, 387,392 lab results, and over 1,505 lab test items. The FEMH Research Ethics Review Committee \footnote{https://www.femhirb.org/} approved the study, and all data were de-identified. We selected patients with at least two recorded visits from each dataset.

For the multi-label classification task in MIMIC-III, we identified the top 10 most prevalent conditions: ``Hypertension, uncomplicated'', ``Cardiac arrhythmias'', ``Fluid and electrolyte disorders'', ``Congestive heart failure'', ``Diabetes w/o chronic complications'', ``Chronic pulmonary disease'', ``Valvular disease'', ``Renal failure'', ``Hypertension, complicated'', and ``Other neurological disorders''. In the FEMH dataset, the top 10 most common diseases include ``Hypertension'', ``Diabetes'', ``Heart disease'', ``Cancer'', ``Cerebrovascular Disease'', ``Kidney Disease'', ``Liver Disease'', ``Asthma'', ``Hyperlipidemia'', and ``Lung Disease''. We applied several established multi-label classification metrics to assess model performance such as Macro-average and weighted-average F1-Scores, precision, recall, and accuracy on the test dataset \cite{lipton2014optimal, hossin2015review, palacio2019evaluation}.



\subsection{Experimental Results}

Table \ref{tab:mimic} and Table \ref{tab:femh} illustrate the training and validation performance of various models, highlighting the effectiveness of our proposed method on the MIMIC-III and FEMH datasets, respectively. In Table \ref{tab:mimic}, our approach outperforms baseline models such as Bert \cite{devlin2018bert}, Mistral-7B-v0.1 \cite{jiang2023mistral}, Llama-2-7B-hf \cite{touvron2023llama}, Meta-Llama2-13B \cite{llama2modelcard}, Meta-Llama3-8B \cite{llama3modelcard}, and Medical-Llama3-8B \cite{medicalllama3modelcard} across all key metrics. Specifically, MEDFuse shows significant improvements over the best-performing LoRA fine-tuned LLM, Medical-Llama3-8B. On the test set, our model performs 1.49\% better in macro F1 score, and similar trends are observed in other metrics. Table \ref{tab:femh} shows MEDFuse consistently outperforms Medical-Llama3-8B on the FEMH dataset. For example, training and validation in precision is a 1.53\% increase, in the recall is a 2.07\% increase, the training accuracy is 0.9311 (0.55\% increase), and validation accuracy is 0.9296 (0.41\% increase). These results validate the robustness and generalizability of our approach, underscoring its potential for accurate and reliable clinical predictions across diverse datasets.

\subsection{Ablation Study}

We conducted an ablation study to examine the contributions of various components in our proposed method, which integrates Medical-Llama3 with a transformer module, utilizing lab tests (LABTEXT) and clinical notes (TEXT). The results highlight clear performance contrast when any component is omitted. Removing both the transformer and LABTEXT results in a 4.81\% drop in training precision and a 4.40\% decrease in validation precision. The most substantial performance reduction occurs when both the transformer and TEXT are excluded, leading to a 29.76\% decrease in training precision and a 30.66\% decrease in validation precision. This underscores the indispensable role of TEXT and the transformer in our method. Even when only TEXT is removed, performance significantly deteriorates, with a 17.14\% decline in training precision and a 14.60\% decline in validation precision. These findings illustrate that each component contributes significantly to the model's overall efficacy. Our full model, combining LLMs and MLTM, demonstrates the highest performance, with a training accuracy of 0.9535 and a validation accuracy of 0.9122.

\section{Conclusion}
In conclusion, we have presented a novel multi-disease diagnostic model that integrates multimodal data, closely mirroring real-life clinical decision-making. By combining fine-tuned LLMs with domain-specific transformers, we achieved enhanced synthesis of structured and unstructured medical data. Using a disentangled transformer further refined this integration, significantly improving disease prediction accuracy. Our experimental results across two practical EHR datasets demonstrated the proposed model's robustness and effectiveness. In future work, we aim to extend our model to cover more complex and rare diseases, enhance its interpretability for clinical use, and evaluate its performance on larger, more varied datasets. We will also explore the integration of real-time and other data modalities \cite{zheng2021effective} to further align our model with dynamic clinical environments.



\bibliographystyle{ACM-Reference-Format}
\bibliography{references.bib}

\end{document}